\documentclass{article}

\usepackage{microtype}
\usepackage{graphicx}
\usepackage{subfigure}
\usepackage{booktabs} 
\usepackage{fancyvrb}
\usepackage{xfrac}

\usepackage{hyperref}

\usepackage{amsmath,amsfonts,bm}

\def\figref#1{figure~\ref{#1}}
\def\Figref#1{Figure~\ref{#1}}

\def\secref#1{section~\ref{#1}}

\def\eqref#1{equation~\ref{#1}}
\def\Eqref#1{Equation~\ref{#1}}

\def\1{\bm{1}}

\DeclareMathAlphabet{\mathsfit}{\encodingdefault}{\sfdefault}{m}{sl}
\SetMathAlphabet{\mathsfit}{bold}{\encodingdefault}{\sfdefault}{bx}{n}

\usepackage[accepted]{icml2021}

\begin{document}

\twocolumn[
\icmltitle{Slower is Better: Revisiting the Forgetting Mechanism in LSTM for Slower Information Decay}



\begin{icmlauthorlist}
\icmlauthor{Hsiang-Yun Sherry Chien}{jhu&intel}
\icmlauthor{Javier S. Turek}{intel}
\icmlauthor{Nicole Beckage}{intel}
\icmlauthor{Vy A. Vo}{intel}
\icmlauthor{Christopher Honey}{jhu}
\icmlauthor{Ted L. Willke}{intel}
\end{icmlauthorlist}

\icmlaffiliation{jhu}{Department of Psychological and Brain Sciences, Johns Hopkins University, Maryland, USA}
\icmlaffiliation{jhu&intel}{Department of Psychological and Brain Sciences, Johns Hopkins University, Maryland, USA; Work done while interning at Intel Labs.}
\icmlaffiliation{intel}{Intel Labs, Hillsboro, Oregon, USA}

\icmlcorrespondingauthor{Hsiang-Yun Sherry Chien}{hsiangyun.chien@gmail.com}
\icmlcorrespondingauthor{Javier S. Turek}{javier.turek@intel.com}

\icmlkeywords{Machine Learning, ICML}

\vskip 0.3in
]



\printAffiliationsAndNotice{}  

\begin{abstract}
Sequential information contains short- to long-range dependencies; however, learning long-timescale information has been a challenge for recurrent neural networks. Despite improvements in long short-term memory networks (LSTMs), the forgetting mechanism results in the exponential decay of information, limiting their capacity to capture long-timescale information. Here, we propose a \emph{power law forget gate}, which instead learns to forget information along a slower power law decay function. Specifically, the new gate learns to control the power law decay factor $p$, allowing the network to adjust the information decay rate according to task demands. Our experiments show that an LSTM with power law forget gates (pLSTM) can effectively capture long-range dependencies beyond hundreds of elements on image classification, language modeling, and categorization tasks, improving performance over the vanilla LSTM. We also inspected the revised forget gate by varying the initialization of $p$, setting $p$ to a fixed value, and ablating cells in the pLSTM network. The results show that the information decay can be controlled by the learnable decay factor $p$, which allows pLSTM to achieve its superior performance. Altogether, we found that LSTM with the proposed forget gate can learn long-term dependencies, outperforming other recurrent networks in multiple domains; such gating mechanism can be integrated into other architectures for improving the learning of long timescale information in recurrent neural networks.

\end{abstract}

\section{Introduction}


Recurrent neural networks (RNNs) are powerful models for capturing the structure of sequence data; they store information from the past in internal hidden states and have recurrent connections to allow these internal states to be combined with input information. These internal representations become ``context" as described in \citealt{elman1990finding}. However, simple RNNs suffer from the vanishing gradient problem, which comes from the backpropagation-through-time (BPTT) algorithm used to train the RNN models. During training through BPTT, the model has difficulty retaining relevant information from previous internal states, especially for events that happened many timesteps in the past \citep{hochreiter1998vanishing}. 
Therefore, the RNN is limited in its ability to learn long-range dependencies (or long-timescale information) of the sequential data. 

One solution to improve long-range information retention
in RNNs is to implement gating mechanisms. By adding gates to the recurrent unit to control the information flow, long short-term memory networks (LSTMs, \citealt{hochreiter1997long}) have successfully improved RNNs' performance in many challenging sequence tasks including language modeling \cite{linzen2016assessing, gulordava2018colorless}. However, information stored in the memory cell in LSTMs decays exponentially  \cite{tallec2018can, mahto2020multi, neil2016phased}, which makes it difficult for LSTMs to learn information over hundreds of timesteps. A simple method to retain information longer in memory cells is to initialize the forget gate bias based on the prior knowledge of sequence lengths \citep{tallec2018can}. The initialization trick not only effectively improves LSTMs' ability to capture information over longer timescales, but also allows one to build a network that can capture information over multiple timescales \citep{mahto2020multi}.  

Although it has been shown that adjusting the forget gate bias can effectively prolong the memory of the LSTMs, 
this requires prior knowledge about the task demands of retaining information the network needs in its internal state. Moreover, such method does not change the fact that information decays exponentially in LSTMs, whereas
real-world information such as natural language, decays following a power law function which decays slower than the exponential function \citep{lin2017critical}. This suggests that a slower forgetting mechanism, such as is seen in power law functions,
might effectively capture information over long timescales. 

In this study, we propose a novel gating mechanism, the power law forget gate, that improves the ability of the LSTM model to learn long-timescale information. This gating mechanism is designed to combat the exponential decay of information by enforcing information decay of the memory cell state to follow a power law function. By empirically testing the performance of LSTM with a power law forget gate (pLSTM) and examining the gating behavior, we find that pLSTM exhibits two major advantages compared to other LSTM models: (1) The decay factor $p$, which controls the rates of information decay, can be flexibly learned to capture a wide range of timescales based on the task demands. This allows the network to learn information over hundreds of timesteps without prior knowledge for initialization. (2) The power law gating mechanism increases the interpretability of model behavior. For example, by examining the unit activation, we discovered that a small set of units with slower decay factors were keeping task-relevant information, yielding a model that is more robust to unit dropout. Altogether, we propose a new gating mechanism that enforces a power law decay of information in the cell state of an LSTM.
We show that this gate 
improves LSTM's performance in tasks that require long-range information. 
We also show how pLSTM achieves its performance by controlling the information decay.    

\section{Background}
\subsection{Related Work}
\label{sec:rel_work}
Many RNN variants were proposed to tackle the vanishing gradient problem and improve long-range information learning. For example, models that use unitary or orthogonal weight matrices and alternative activation functions to prevent gradient vanishing during BPTT \cite{arjovsky2016unitary,kerg2019non,lezcano2019cheap, jing2017tunable}; models that use alternative learning methods such as real-time recurrent learning to avoid tracing the gradient back in time \citep{benzing2019optimal, schmidhuber1992fixed, mujika2018approximating}; models that add additional memory cells to store information \citep{chandar2019towards, voelker2019legendre, mikolov2015learning}, models that are initialized based on prior knowledge \citep{van2018unreasonable, tallec2018can}, and models that revisited the gating mechanisms of gated RNNs \citep{gu2020improving,neil2016phased, jing2019gated}. 

Our work is closely related to prior studies that modified the gating mechanisms of RNNs. For example, \citet{neil2016phased} proposed the phasedLSTM, where a ``time gate" is added 
to control the periodic update of the memory cell.
The time gate improves the temporal precision of the network, and allows the network to learn irregularly sampled data. When setting a longer period in the time gate, the network also showed advantage in learning long sequences.

Most recently, \citet{gu2020improving} proposed a gating mechanism, the refine gate, by implementing an adjustment function to the original forget gate activation to capture information over longer timescales than the original forget gate. An additional gating parameter is added for the network to learn how to adjust the activation curves according to different tasks. The authors also proposed a uniform initialization method modified from the initialization method proposed by \citet{tallec2018can}, which allows the network to capture a wide range of timescales. 

The current study shares a similar strategy with these prior studies by modifying the gating mechanism to prolong the memory of RNNs; however, our method differs from prior studies by directly controlling the information decay in LSTMs to follow a specific function, in our case, the power law function. This choice of decay is directly
motivated by the nature of real-world information decay \citep{lin2017critical}. The power law gated LSTM covers the advantages shown in prior studies: The network can flexibly learn information over a wide range of timescales, and has better temporal precision with time as an input to the gate. In the meantime, parameters such as decay factor and reference time derived from the power law function are more interpretable compared to other LSTM variants.

\subsection{Information Decay in LSTMs}
Gated RNNs introduce a gating mechanism that controls the flow of information and overcomes vanishing gradients during backpropagation. A gate functions as a switch that controls the amount of information that is allowed to pass through it. LSTMs \citep{hochreiter1997long} and Gated Recurrent Units (GRU) \citep{cho2014learning} are examples of recurrent models that exploit gating mechanisms to support the retention of information for longer inputs. In particular, the LSTM defines an input gate $i_t$, a forget gate $f_t$, a cell state update gate $\tilde{c}_t$, and an output gate $o_t$:
\begin{eqnarray}
f_{t} &=& \sigma\left(U_{f}x_{t}+W_{f}h_{t-1}+b_{f}\right) \nonumber\\
i_{t} &=& \sigma\left(U_{i}x_{t}+W_{i}h_{t-1}+b_{i}\right) \nonumber\\
o_{t} &=& \sigma\left(U_{o}x_{t}+W_{o}h_{t-1}+b_{o}\right) \nonumber\\
\tilde{c}_{t} &=& \tanh\left(U_{c}x_{t}+W_{c}h_{t-1}+b_{c}\right) \nonumber\\
c_{t} &=& f_{t} \odot c_{t-1} + i_{t} \odot \tilde{c}_{t}  \label{eq:lstm} \\
h_{t} &=& o_{t} \odot \tanh \left( c_t \right), \nonumber 
\end{eqnarray}
where $\sigma(\cdot)$ is the sigmoid function, and $\odot$ stands for the Hadamard product. The hidden state $h_t$ tracks the model dynamics and partially controls the activation of the gates, whereas the memory cell state $c_t$ stores information.

Although LSTMs have been prominent at processing longer sequences than RNNs, \citet{mahto2020multi} suggests that information in the memory cell state decays at an exponential rate. Their analysis is conducted in a ``free regime'' case where there is no external input to the network after timestep $t_0$, i.e., $x_t=0$ for $t>t_0$. Information leakage through the hidden state is ignored, assuming $W_c=0$, $W_f=0$, and $b_c=0$. Therefore, the cell state update in \Eqref{eq:lstm} becomes
\begin{equation}
c_{t}  = f_{t}\odot c_{t-1} = c_0\odot e^{\log \left( f_0 \right) \left(t-t_0\right)},
\label{eq:exprecur}
\end{equation}
where $f_0=\sigma\left(b_f\right)$ and $c_0$ is the cell state at timestep $t_0$. \Eqref{eq:exprecur} shows that the cell state follows an exponential decay with a $\log\left(f_0\right)$ factor under such conditions. Consequently, the forget gate bias term, $b_f$, controls the rate of decay. A high bias value $b_f$ would yield a slower decay rate. This fact was previously observed by \citet{tallec2018can}, who derived the ``chrono initialization'' method, which enables LSTMs 
to learn to capture information for longer sequences. This method proposes to initialize the forget gate bias as $b_{f}\sim \log(\mathcal{U}([1, T_{max}-1]))$, and the input gate by $b_{i} = -b_{f}$.
A caveat of the chrono method is that the maximum dependency range $T_{max}$ \emph{should be known a priori}.

\section{Power Law Forget Gate}
\label{sec:powerlawgate}
\subsection{Slower Information Decay}

\begin{figure}[tb]
\vskip 0.2in
\begin{center}
\centerline{\includegraphics[width=\columnwidth]{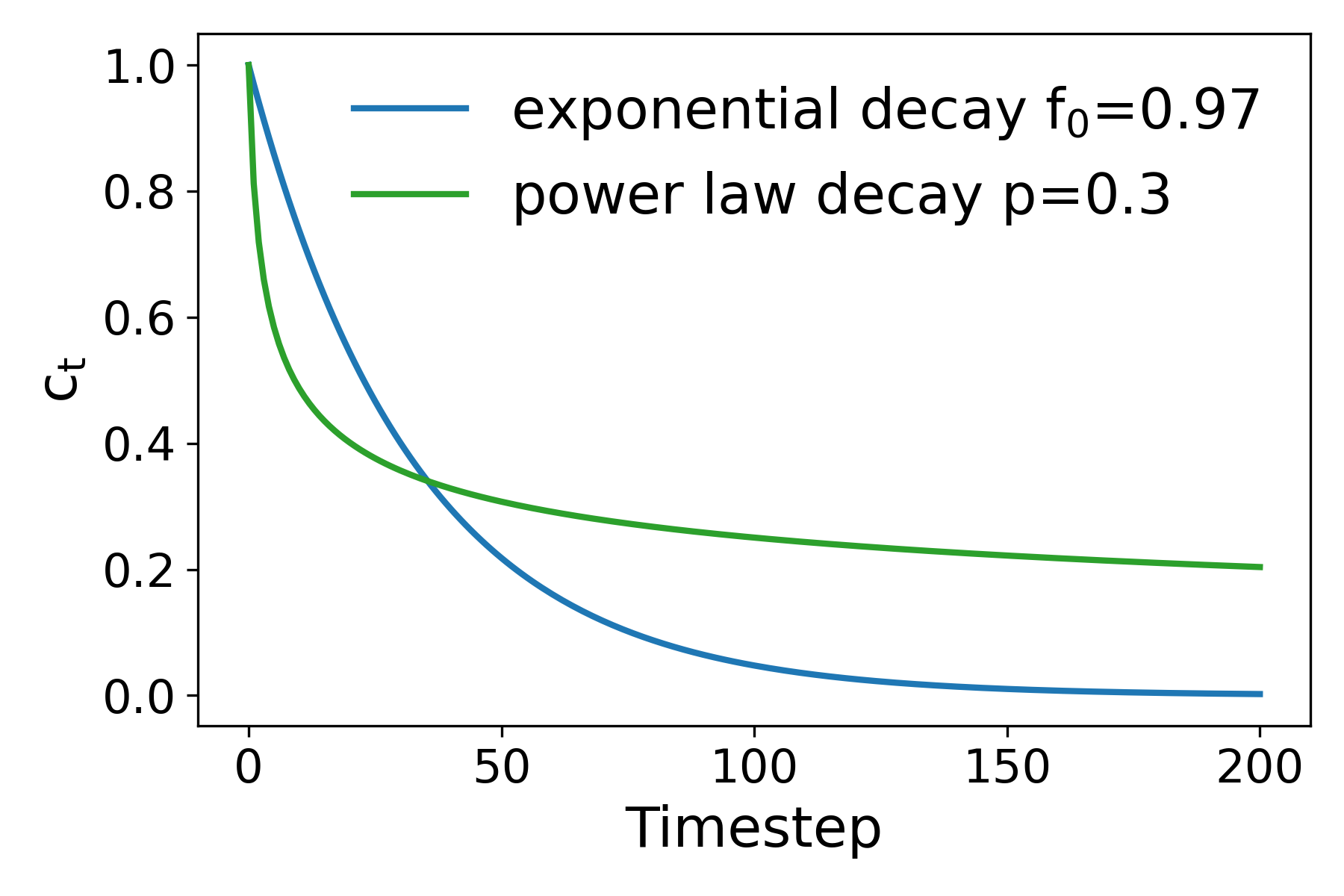}}
\caption{\textbf{Simulation of internal state decay following an exponential and a power law function} In LSTM, the cell state decay follows the exponential recursion, $c_{t}=f_0 \odot c_{t-1}$; while in pLSTM, the cell state decay follows the power law recursion, $c_{t}= {(\frac{t-t_0+1}{t-t_0})}^{-p}\odot c_{t-1}$. Internal state decays much slower following the power law function than the exponential function.}
\label{decaysimulation}
\end{center}
\vskip -0.3in
\end{figure}

The exponential decay due to the forget gate in \Eqref{eq:exprecur} poses a limitation to capturing information for long-term dependencies. \Figref{decaysimulation} shows a decay simulation with decay rate $f_0=0.97$ over 200 timesteps. 
The exponential decay (blue) exhibits a decay that tends very fast to 0.0 within the given time range. Of course, increasing $f_0$ would solve the problem. It then seems worth asking: would it be possible to have an alternative gate with a slower information decay than the exponential decay of the LSTM forget gate?

We examined the possibility of information decay along a power law. As noted above, several real-world phenomena generate data where the temporal dependencies decay along a power law \cite{lin2017critical}. A power law function of the form $(t-t_0+1)^{-p}$ can grant us a slower decay rate.
In \figref{decaysimulation}, we see that power law decay (green) with $p=0.3$ shows a slower decay that keeps information longer, beyond the 200 timesteps. 

The slower decay rate of a power law function satisfies our objective. Next, we derive a power law-based forget gate to improve over the exponential decay. As a starting point, consider the ``free regime'' in \Eqref{eq:exprecur}. The parallel version with a power law function is:
\begin{eqnarray}
c_{t} &=& c_0 \odot (t-t_0+1)^{-p} \nonumber \\
&=&  \left(\frac{t-t_0+1}{t-t_0}\right)^{-p} \odot c_{t-1},
\label{eq:plrecur}
\end{eqnarray}
where $t_0$ is a reference time point indicating the start of information decay and $c_0$ is the cell state value at that moment.
\Eqref{eq:plrecur} has a recurrent coefficient that is smaller than 1 (due to the negative sign in the power), suggesting that any positive power $p>0$ is acceptable. However, the power law decay is slower as $p$ approaches $0$, suggesting that long-range dependencies would be better captured with small values of $p$. Unlike its exponential counterpart, the power law recurrence \Eqref{eq:plrecur} depends both on the timestep $t$ and the reference timestep $t_0$. 

\begin{figure*}[t]
\vskip 0.2in
\begin{center}
\centerline{\includegraphics[trim=0 7 10 10, clip,width=\textwidth]{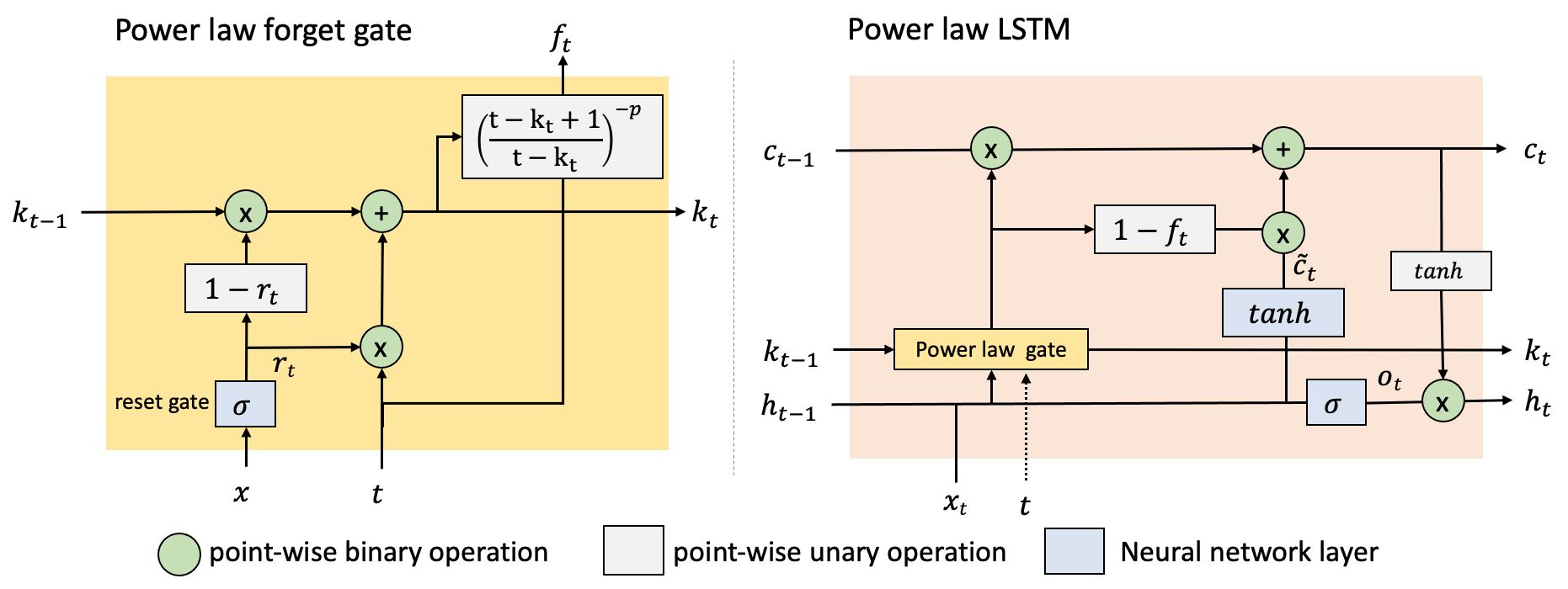}}
\caption{\textbf{Power law forget gate implemented in an LSTM.} (Left) The power law forget gate takes current time $t$ and prior reference time $k_{t-1}$ as input. After the input $x_t$ activates the reset gate $r_t$, the reference time $k_t$ is calculated. Then the final forget gate value, $f_t$, is calculated using $t$, $k_t$ and the power parameter $p$. (Right) The LSTM with the power law forget gate $f_t$ used in the current study. Most LSTM architecture is preserved, except that the input gate $i_t$ is replaced with $1-f_t$.}
\label{pLSTM}
\end{center}
\vskip -0.2in
\end{figure*}

\subsection{Controlling the Reference Time}
The main idea of the new power law gating mechanism is to replace the forget gate $f_t$ in \Eqref{eq:exprecur} with a derived power law gate based on the recursive \Eqref{eq:plrecur}. 
When new information is to be captured by the recurrent network at timestep $t$, the forget gate $f_t$ is expected to be 0. Namely, we need to control the reference time $t_0$ for this objective, setting it equal to $t$. This implies that we should be able to modify $t_0$. Therefore, we introduce a new variable, $k_t$, that keeps a reference time $t_0$ and can be updated by the network when needed:
\begin{eqnarray}
c_{t} &=& {\left(\frac{t-k_t+1}{t-k_t}\right)}^{-p} \odot c_{t-1} \label{eq:cpower} \\ 
k_t &=& r_t\odot t + (1-r_t)\odot k_{t-1}. \label{eq:ref_time}
\end{eqnarray}
The variable $k_t$ is defined recursively, hence, adding to the cell memory and the hidden states of an LSTM. Further, we propose to initialize it to $k_0=0$ (as $t\ge 1$).
\Eqref{eq:ref_time} incorporates a reset gate $r_t$ that controls when the reference time $k_t$ should be updated based on the input and the recurrent information. We define the reset gate as 
\begin{equation}
    r_t = \sigma\left(U_r x_t + W_r h_{t-1} + b_r\right), 
    \label{eq:reset_gate}
\end{equation} 
where $U_r, W_r$ are the weight parameters, and $b_r$ the bias parameters.
Specifically, when the reset gate is activated $r_t \to 1$, 
the reference time $k_t$ would be reset to the current time point $t$; otherwise, $k_t$ is kept as the prior reference time $k_{t-1}$ to allow information to decay.

Equations \ref{eq:cpower}-\ref{eq:reset_gate} form the basis of our power law-based gating mechanism. Similar to the LSTM, we extend \Eqref{eq:cpower} to include the cell state update gate output $\tilde c_{t}$.
\begin{eqnarray}
f_{t} &=& \left(\frac{t-k_t+1}{t-k_t}\right)^{-p} \label{eq:forget_gate_plstm}\\
c_{t}&=& f_{t}\odot c_{t-1}+i_t\odot\tilde c_{t}. \label{eq:cell_state_plstm}
\end{eqnarray}
In general, we consider the input gate $i_t$ to be a separate gate as in the LSTM. However, in practice it is possible to replace the input gate by $i_t=1-f_t$, a modification suggested by \citet{van2018unreasonable}. The argument is that keeping and removing previous information in the cell state $c_t$ should be coupled. 
Replacing $i_t$ with $1-f_t$ yields $\sim\sfrac{3}{4}$ of total parameters in pLSTM compared to LSTM when the hidden sizes are equal. 

In sum, we define the power law gated LSTM (pLSTM) recurrent cell based on the LSTM definition but with the forget gate $f_t$ derived from Equations \ref{eq:ref_time}-\ref{eq:cell_state_plstm}.


\subsection{Backpropagating Through the Power Law Gate}
\label{sec:gradient}
The power law gate follows the same principles as the LSTM gates, thus avoiding vanishing gradients during backpropagation through time. Nevertheless, we derive the gradient with respect to the reset gate in \Eqref{eq:reset_gate} to investigate the effect of the new forget gate. Given the loss function $L$, we compute the gradient with respect to $r_t$:
\begin{equation*}
\frac{\partial L} {\partial r_t} = \frac{\partial L}{\partial c_t}\cdot\frac{\partial c_t}{\partial f_t}\cdot\frac{\partial f_t}{\partial k_t}\cdot\frac{\partial k_t}{\partial r_t}.
\end{equation*}
From this derivation, we note that the term $\frac{\partial f_t}{\partial k_t}$ holds the key difference and the complex functionality of the gate.
The gate derivative term is then given by
\begin{equation}
\frac{\partial f_t}{\partial k_t} = -p\odot \frac{\left(t-k_t\right)^{p-1}}{\left(t-k_t+1\right)^{p+1}}.
\label{eq:grad_forget_gate}
\end{equation}
In the case that the power $p \in (0,1)$, 
the partial derivative  $\sfrac{\partial f_t}{\partial k_t} \to \infty$. This numerical instability makes the backpropagation through time unstable. To overcome this, we add an $\varepsilon$ term in \Eqref{eq:forget_gate_plstm} such that $f_t=\left(\frac{t-k_t+1}{t-k_t+\varepsilon}\right)^{-p}$. $\varepsilon$ should be set to a small value. In all the experiments reported here, we set $\varepsilon=0.001$.

\subsection{Learnable Power Parameter}
The power parameter $p$ in \Eqref{eq:forget_gate_plstm} determines the decay rate in the power law. A small $p$ would make the information to decay slower compared to a large $p$. Keeping information over hundreds of timesteps would likely require $p < 1$ (see Section \ref{sec:eval_p}). Clearly, selecting the right value of $p$ is important. One possibility is to manually set the value of $p$ based on prior knowledge of the task demands, similar to the chrono initialization method \citep{tallec2018can}. 

Alternatively, we consider the power $p$ as a learnable parameter of the current model. This allows each unit in the network to learn the best rate of decay. Given that the range $(0,1)$ is the most useful for the power $p$, we propose to compute $p$ as $p=\sigma(\hat{p})$, and allow $\hat{p}$ to be any real value. If not specified, in all of our experiments, we initialize the values $\hat{p}$ such that the values of $p$ are uniformly distributed on the $(0,1)$ range, i.e., $p \sim \mathcal{U}(0,1)$.



\section{Experimental Results}
\label{sec:results}

We tested the new power law forget gate incorporated into an LSTM on several tasks that require retaining information over long timescales, such as a copy memory task. We compared the performance of the power law LSTM (pLSTM) to a vanilla LSTM and other gated recurrent models. All experiments were implemented using Pytorch 1.7 \citep{pytorch}. Unless otherwise specified, the models are trained with Adam \citep{kingma2017adam} at a learning rate of $0.001$, $\beta_1=0.9$, and $\beta_2=0.999$. 

\begin{figure*}[t]
\vskip 0.2in
\begin{center}
\centerline{\includegraphics[width=\columnwidth*2]{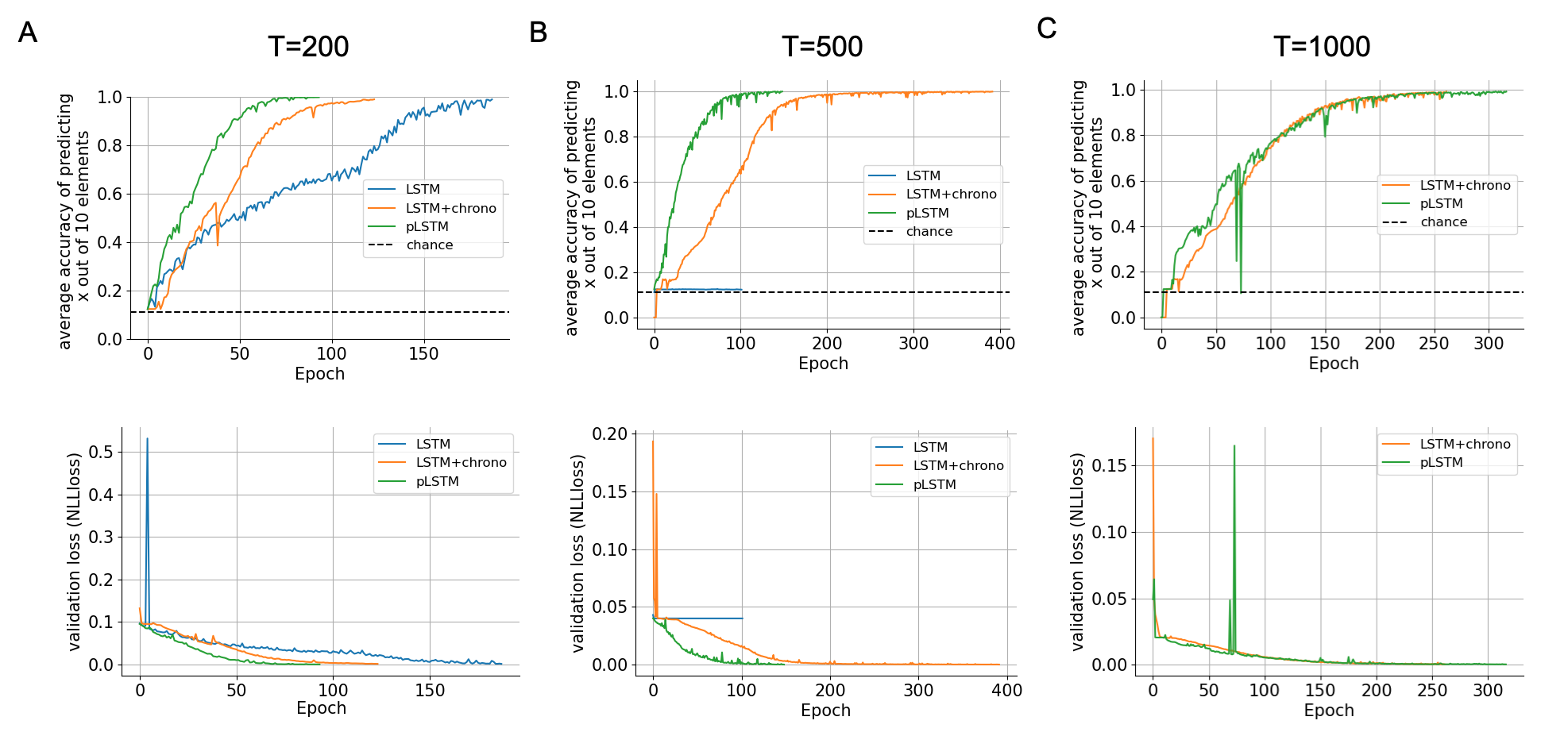}}
\caption{\textbf{Accuracy and validation loss on the copy task.} \textbf{A.} At $T=200$, the pLSTM converges faster than the LSTM with and without chrono initialization. \textbf{B.} At $T=500$, the pLSTM converged faster than LSTM-chrono. The vanilla LSTM failed to learn the task. \textbf{C.} At $T=1000$, the pLSTM and LSTM-chrono both learned the task, with a similar rate of convergence.}
\label{copytask}
\end{center}
\vskip -0.3in
\end{figure*}

\subsection{Copy Memory Task}
\label{sec:copytask}
To examine whether the power law forget gate in the pLSTM does in fact result in information decay at a slower rate, we compare the performance of LSTM, LSTM with chrono initialization (LSTM-chrono) \citep{tallec2018can}, and the pLSTM on the copy memory task \citep{hochreiter1997long}. The copy task was designed to examine the networks' ability to memorize information for a certain amount of time, $T$. 
The dataset consists of symbols $\{a_{i}\}_{i=0}^{m+1}$: the first $m$ elements $a_{0}, a_{1}, ..., a_{m-1}$ are the ``target symbols'', $a_{m}$ is the ``dummy symbol'', and $a_{m+1}$ is the ``signal symbol''. The first $n$ elements of the input sequence contain the randomly sampled target symbols, followed by $T$ dummy symbols, one signal symbol, and $n-1$ dummy symbols. The target sequence contains $n+T$ dummy symbols, followed by $n$ target elements. To achieve the task, the model needs to memorize the first $n$ elements, and after keeping those elements for $T$ time steps, output those elements in order once it sees the signal symbol. In our experiments, $m=8$ and $n=10$, and we examined the models' ability to retain information over $T=200,$ $500$ and $1000$ timesteps.

For all the experiments, we set batch size to 128, hidden unit size to 128, and learning rate to 0.001, following prior studies \citep{arjovsky2016unitary,tallec2018can}. Optimization was performed using RMSprop \citep{tieleman2012lecture} with a moving average parameter of 0.9. The training dataset contains 100K sequences, and the validation dataset 10K sequences. 
We define task accuracy as the proportion of successfully reconstructed elements out of the 10 target elements (Figure \ref{copytask}, top row).

For the LSTM-chrono, we initialized the LSTM forget gate bias with $T_{max} = \sfrac{3T}{2}$, the parameter chosen by \citet{tallec2018can} for their copy task experiment. 
The pLSTM was able to learn the copy task in all conditions. It also converged faster than LSTM-chrono for $T=200$ and $500$. This suggests that the power law gate helps pLSTM to retain information over long-timescales, with no need for prior knowledge of task demands to initialize the network. 

In addition to the copy task with fixed $T$, we also tested the models' performance on the ``various copy task" where $T$ varies from trial to trial. We found that pLSTM with uniform initialization was able to learn the various copy task with a small subset of units keeping information until the reproduction starts (Supplemental material, Figure \ref{fig:copytask-varyT}). We also found that pLSTM is able to generalize the task to different $T$ while training with a fixed $T$ (Supplemental material, Figure \ref{fig:copytask-generalization}).  


\subsection{Sequential MNIST}
Next, we examine the performance of the pLSTM on the sequential MNIST classification task and the permuted sequential MNIST (pMNIST) \cite{le2015simple}. In this task, a digit image is converted to a 784-length sequence of pixels by reading it row after row from top to bottom. In pMNIST, the element order in the sequences is permuted. We partition the MNIST training set into 50K sequences for training and 10K for validation. The test set contains 10K sequences.

We evaluated the LSTM and pLSTM architectures with hidden sizes of 256 and 512 units for 3 different seeds each. As noted by \citet{pmlr-v119-gu20a}, training LSTMs with more hidden units can be unstable and the model failed to converge with some seeds, therefore we report on three successful ones. We report in Table \ref{tab:MNIST} the average performance of the models, and compare them with other recurrent networks from previous works in Section \ref{sec:rel_work}. The results show that the pLSTM captures long-range information better than the LSTM and other recurrent architectures. Unlike the LSTM, the pLSTM was able to successfully train every run.

\begin{table}[tb]
\caption{Test accuracy for pixel-by-pixel digit classification for LSTM and power law LSTM (pLSTM). Average results for successfully trained models over 3 random seeds. Results from previous works are shown for reference.}
\label{tab:MNIST}
\vskip 0.15in
\begin{center}
\begin{small}
\begin{sc}
\begin{tabular}{lcc}
\toprule
Method & MNIST & pMNIST \\
\midrule
LSTM 256 (ours) & 98.7\% & 91.3\% \\
LSTM 512 (ours) & 98.6\% & 91.7\% \\
pLSTM 256 &  \textbf{99.1\%} & 94.4\% \\
pLSTM 512 &  \textbf{99.1\%} & \textbf{95.6\%} \\
\midrule
ExpRNN & 98.7\% & 94.7\% \\ 
JANET  & 99.0\% & 92.5\% \\ 
NRU  & -- & 95.4\% \\ 
SRU & -- & 92.5\% \\
URNN & 97.6\% & 94.5\% \\
LSTM & 91.9\% & 91.8\% \\
\bottomrule
\end{tabular}
\end{sc}
\end{small}
\end{center}
\vskip -0.1in
\end{table}

\subsection{Character-Level Language Modeling}
Beyond synthetic datasets, we also conducted experiments on real-world language datasets, which have been shown to exhibit power law decay of long-range information \citep{lin2017critical}. We considered two language corpora: Penn Treebank Corpus (PTB, \citealt{marcus-etal-1993-building}) and Text8\footnote{\url{http://mattmahoney.net/dc/textdata.html}}. 
For experiments using the PTB dataset, the pLSTMs and LSTMs have $\sim$2.15M  parameters to enable comparison with previous models \citep{chandar2019towards}. This results in 716 and 828 units for the LSTMs and pLSTMs, respectively.

We trained the model using sequence lengths 150 or 500 and a batch size of 128. We evaluated the models’ performance by measuring bits-per-character (BPC) and trained them for a total of 50 epochs. We measured test BPC of the epoch with the best validation BPC. We report results averaged across 5 random seeds (Table \ref{tab:charLM}). For both sequence lengths, the pLSTM achieved a small improvement over the LSTM, and outperformed other models with the same number of parameters (Table \ref{tab:charLM}). Note that our LSTM achieved a lower BPC than \citet{chandar2019towards}, but they only trained for 20 epochs. For Text8, a much larger dataset, we trained bigger models ($\sim3.22$M total parameters) with 2 layers. Here the pLSTM also slightly outperformed the LSTM on the test BPC. 

\begin{table}[tb]
\caption{Test BPC for character-level language modeling.}
\label{tab:charLM}
\vskip 0.1in
\begin{center}
\begin{small}
\begin{sc}
\begin{tabular}{lrcc}
\toprule
Method  & Dataset  & Test BPC \\
\midrule
LSTM (ours)  &PTB  & 1.426 \\
LSTM-500 (ours) &PTB  & 1.403 \\
pLSTM &PTB  & \textbf{1.420} \\
pLSTM-500 &PTB  & \textbf{1.396} \\
\midrule
LSTM     &PTB  & 1.48 \\
LSTM-chrono    &PTB  & 1.51 \\
GRU     &PTB  & 1.45 \\
JANET    &PTB & 1.48 \\ 
GORU     &PTB  & 1.53 \\
EURNN  &PTB  & 1.77 \\ 
NRU  &PTB  & 1.47 \\ 
\midrule
LSTM-500  &Text8  &1.552\\
pLSTM-500 &Text8  & \textbf{1.544} \\
\bottomrule
\end{tabular}
\end{sc}
\end{small}
\end{center}
\vskip -0.1in
\end{table}


\subsection{IMDB Sentiment Classification}
We also looked at the performance of the pLSTM on a sentiment classification task. In the IMDB dataset, each review is classified as either positive or negative \citep{maas2011}. The median review length is 202 words (minimum 5, maximum 2789). We pad or truncate the sequences to achieve a fixed length of 400 words. We set aside a random 10\% of the training set (2500 of 25K examples) for validation. After training for 20 epochs, we take the epoch with the lowest validation loss and report the accuracy on the test set (25K examples) for that epoch. Following previous work, we used a 100d pretrained GLoVe embedding with a dictionary size of 25K \citep{ruschMishra2020}.
Since we froze the embedding weights, the number of parameters we report excludes them. Each model is a single layer LSTM with an approximately matched number of parameters ($\sim$118K parameters; 128 units for vanilla LSTM and LSTM with chrono; 154 units for pLSTM). We applied dropout (rate 0.2) to the output of the LSTM. The learning rate was optimized separately for each model by a hyperparameter grid search (LSTM and LSTM with chrono: $0.001$; pLSTM: $0.0005$). We report results averaged across 3 random seeds.

\begin{figure}[tb]
\vskip 0.2in
\includegraphics[width=\columnwidth]{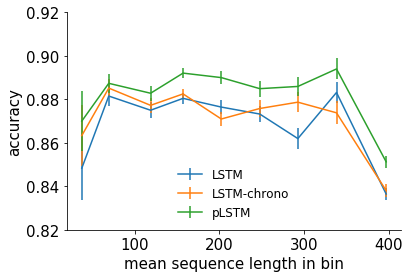}
\caption{\textbf{IMDB sentiment classification accuracy on different sequence lengths.} The pLSTM model shows improved task performance for sequences in the test set that are longer than 150 words. Values shown are the mean and standard error over all sequences across all 3 random seeds within a given bin.}
\label{fig:imdb}
\vskip -0.2in
\end{figure}

The pLSTM achieves a higher accuracy (88.1\%) over the LSTM (86.8\%) and LSTM with chrono (87.0\%). Further analysis shows that the pLSTM selectively improves performance for longer sequence lengths (Figure \ref{fig:imdb}). Overall, the experiments on language datasets suggest that the pLSTM can effectively capture information present in naturalistic data, especially in longer sequences.

\begin{figure}[t]
\vskip 0.2in
\begin{center}
\centerline{\includegraphics[width=\columnwidth]{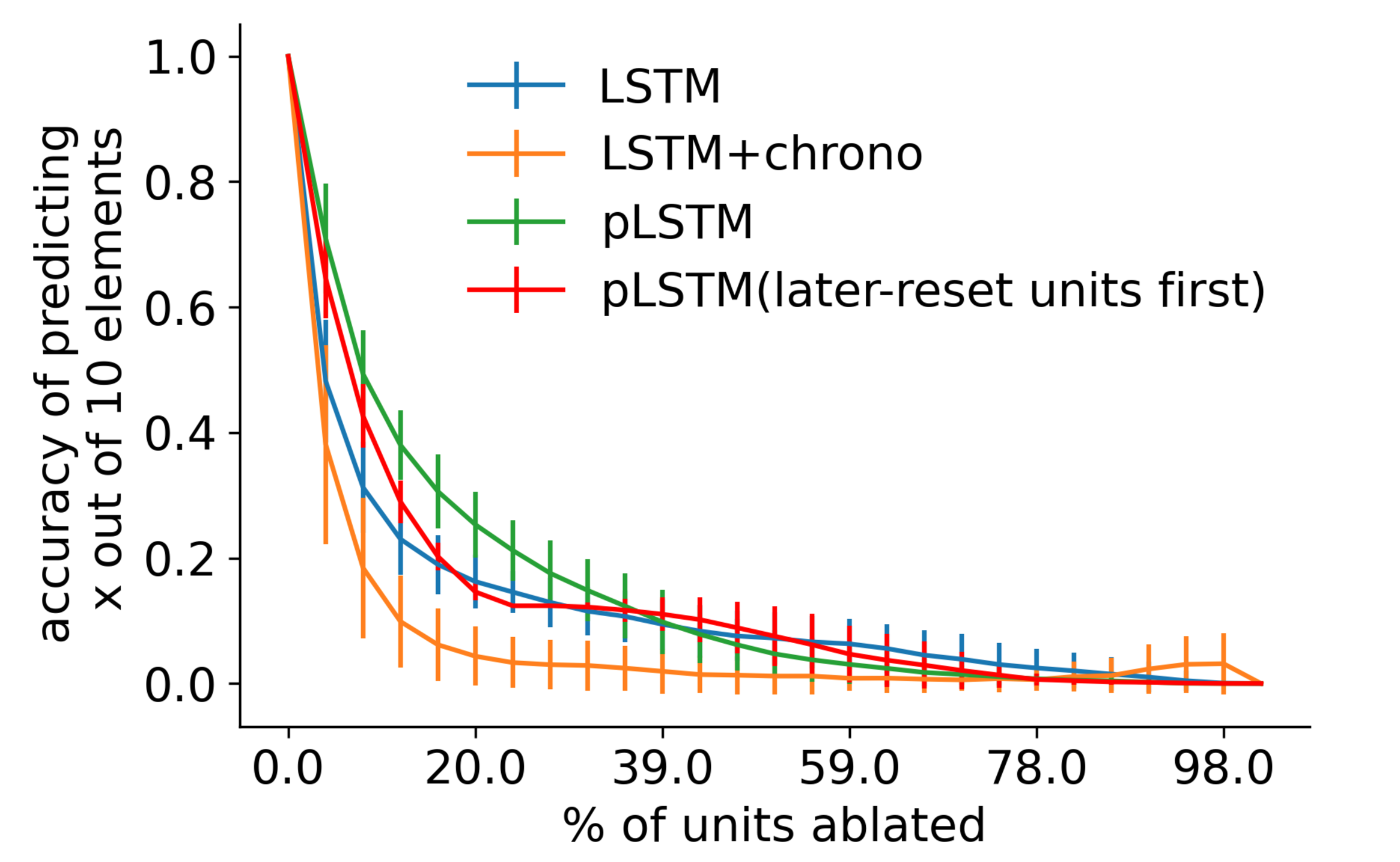}}
\caption{\textbf{The pLSTM is robust to random unit ablation.} We plot accuracy on the copy task ($T=200$) as 5 units are ablated at a time ($\sim 4\%$ of total), for a total of 100 ablated units. The accuracy dropped more slowly in the pLSTM. This indicates that the pLSTM uses a sparse, local representation over relatively few units to solve the task.  Moreover, when ablating the critical later-reset units first, the performance dropped significantly faster.}
\label{ablation}
\end{center}
\vskip -0.3in
\end{figure}

\section{Characteristics of the Power Law LSTM}
\subsection{Effect of Decay Factor $p$} 
\label{sec:eval_p}
Does information decay in pLSTM really follow the power law function, as we expect? If so, the decay factor $p$ should determine the amount of information retained by the network, and thus affect the model's task performance. To examine this, we tested the performance of the pLSTM on the copy task described above ($T=200$), except we fixed $p$ to different values each time. 
We found that the decay factor $p$ indeed affects the pLSTMs' convergence speed on the copy task (Figure \ref{fig:copytask-varyp}): models with smaller $p$ converged faster than models with larger $p$, while models with $p \geq 1.0$ had difficulty converging after training for 1000 epochs.

\subsection{Reset Units and Learned $p$ Distribution}
\label{sec:kt&pdist}
In this section we address two questions: (1) How do individual units in the pLSTM retain long-range information? and (2) how does the learned $p$ distribution change for tasks that require the network to keep information for different lengths? 
To answer the first question, we visualized how the reference time $k_t$, controlled by the reset gate $r_t$, was learned in each individual unit during the copy task (\secref{sec:copytask}). We found that among the 128 units, around $\sfrac{1}{5}$ of the units are critical for retaining information -- these units do not reset $k_t$ until the reproduction of target elements starts (``later-reset units"). In contrast, the majority of units were resetting constantly to keep $k_t$ close to current time $t$ (Figure \ref{fig:pdist}). Interestingly, the proportion of ``later-reset units" in the $T=200$ and $500$ conditions are similar, indicating that in the pLSTM, long-range information extended to hundreds of time steps can be learned by a small subset of units.

Next, we examined the learned $p$ distribution after the model reaches over $99\%$ accuracy on the copy task. If $p$ controls how much the information is retained, then we should see a different $p$ distribution in the $T=200$ vs. the $T=500$ condition. We found that as expected, $p$ learned in the  
$T=200$ condition is significantly larger than $p$ learned in the $T=500$ condition ($\mu _{T200}=0.41,\,\mu _{T500}=0.19,\, t=6.9,\, p<0.001$, Cohen's d=0.87). Moreover, the decay factor $p$ of the ``later-reset units" (marked in orange in Figure A.4) are relatively small in the distribution, indicating that units which retain information for longer are associated with a slower decay rate. Altogether, these observations are consistent with our hypothesis that information decay is modulated by the decay factor of the power law forget gate.

\subsection{Ablation Analysis}
We tested model robustness by measuring the model performance after random unit ablation. Since we observed that a small subset of units retain information for the whole memorization period $T$, we hypothesized that the pLSTM implements a local representation strategy for retaining information over long-timescales. Thus its performance should be more robust to random unit ablation as compared to LSTMs. We randomly sampled 100 of the 128 units and  ablated 5 units at a time. The results showed that the pLSTM is indeed more robust to unit ablation: when randomly ablating units in the network, the pLSTM showed slower decay of task performance compared to LSTM and LSTM-chrono in both $T=200$ and $T=500$ conditions (Figure \ref{ablation}, Figure A.5). Interestingly, we found that LSTM with chrono initialization was more susceptible to the unit ablation than both LSTM and pLSTM in the $T=200$. This suggests that chrono initialization leads to a distributed representation for solving the task and thus is fragile to ablation. 

To examine whether the ``later-reset units" are particularly important for the pLSTM performance than other units, we performed another experiment that ablated the later-reset units first. 
The results showed that ablating the later-reset units causes the accuracy to drop more than ablating random units (Figure \ref{ablation}), providing further evidence that these units are critical to retaining information over long timescales.

\subsection{Time Interval Information in the pLSTM}
\label{sec:FDT}
Finally, we examined how pLSTM can use its information about time from the power law gate 
in a task that requires temporal precision. In the frequency discrimination task, each network must learn to classify different sine wave sequences that are sampled either at regular $\Delta t$ intervals (synchronous conditions, Table \ref{tab:FD}) or at variable intervals (asynchronous condition) (\citealt{neil2016phased}, see supplemental materials for experiment details). The LSTM struggles to perform when the sampling is performed very frequently ($\Delta t$=0.1) or when it is variable. 
In the supplement, we show a derivation for \Eqref{eq:cpower} that allows the model to incorporate a non-unit, $\Delta t$ interval, between timesteps, rather than assuming the time elapsed is 1. This is a unique property of the pLSTM which could improve its performance over the LSTM.

\begin{table}[bt]
\caption{Classification performance on the frequency discrimination task when the current timestamp is included as input (``{\sc w/time}'') versus when time is excluded.}
\label{tab:FD}
\vskip 0.15in
\begin{center}
\begin{small}
\begin{sc}
\begin{tabular}{lccc}
\toprule
 & \shortstack{sync. \\ $\Delta t = 1$} & \shortstack{sync. \\ $\Delta t = 0.1$} & async \\
\midrule
 LSTM w/time    & $99.45(0.2)$ & $70.7(17.9)$  & $82.73(0.8)$ \\
 Chrono w/time  & $99.45(0.1)$ & $98.47(0.6)$  & $86.25(1.4)$ \\
 pLSTM w/time & $99.49(0.1)$ & $99.41(0.5)$  & $84.64(2.1)$ \\
 \midrule
 LSTM  & $99.37(0.1)$ & $71.7(22.2)$  & $68.59(1.4)$\\
 Chrono   & $99.29(0.3)$ & $99.02(0.4)$  & $69.64(0.8)$ \\
 pLSTM  & $99.29(0.1)$ & $99.15(0.4)$  & $92.57(0.5)$ \\
\bottomrule
\end{tabular}
\end{sc}
\end{small}
\end{center}
\vskip -0.1in
\end{table}

 
In the asynchronous case, we find that the LSTM and LSTM-chrono perform well only when timing information is passed as an input (accuracy $\approx 83$, 86 respectively). The pLSTM performs well whether or not timing information is provided to the network (accuracy $\approx 84$, Table \ref{tab:FD}). This suggests that the gating mechanism learns and maintains information related to time within the gate itself and can use this information for prediction.

\section{Conclusion}

In this study, we proposed a novel gating mechanism, the power law forget gate. When implemented in an LSTM, called pLSTM, the new forget gate effectively results in the decay of cell state information via a power law function.
This type of slow decay allows the network to learn long-range information over hundreds of timesteps. We empirically tested the performance of the pLSTM 
tasks requiring long-timescale information such as the copy task, image classification, and with naturalistic 
language datasets. We show 
that the pLSTM can effectively capture information over a wide range of timescales by adjusting the decay factor in the power law forget gate. Further analyses showed that the pLSTM learned to perform the copy task using only a small subset of units. This allows
the model to be robust to unit dropout. Further, the learnable decay factor results in a model that can flexibly learn even when relevant task information is not available at the outset.
Finally, even without 
time as an input, the pLSTM also showed an advantage on learning a task that requires temporal precision, suggesting that the pLSTM gate learns information about time which may further increase its capacity to capture long range time dependencies as well as time dependencies at multiple scales.

\bibliography{references}
\bibliographystyle{icml2021}

\newpage

\twocolumn[
\icmltitle{Supplementary Materials for Slower is Better: Revisiting the Forgetting Mechanism in LSTM for Slower Information Decay}

\vskip 0.3in
]
\counterwithin{figure}{section}
\renewcommand{\thefigure}{A.\arabic{figure}}
\setcounter{figure}{0} 

\appendix

\section{Variable Time Points}
\subsection{Model Derivation for variable time}
\label{sec:pLSTM_variable}
Here we address the time information more formally.
In the main paper we showed the derivation of the power law gate for unit timesteps. Let us recall the recurrent formula for the unit timestep case:
\begin{equation}
c_{t} = {\left(\frac{t-t_0+1}{t-t_0}\right)}^{-p} \odot c_{t-1}.
\label{eq:plrecur_supp}
\end{equation}
Sometimes, it could make sense to have variable timesteps, such as non-uniform sampling of time-series \cite{neil2016phased}. Dealing with such cases, requires a time reference to be given in the input. Let us assume that the temporal difference between two time points is $\Delta t$. Therefore, \Eqref{eq:plrecur_supp} becomes
\begin{equation}
c_{t} = {\left(\frac{t-t_0+1}{t-\Delta t-t_0+1}\right)}^{-p} \odot c_{t-\Delta t},
\label{eq:plrecur-deltat}
\end{equation}
where $c_{t-\Delta t}$ is the cell state at the previous time point $t-\Delta t$. The change in the recurrent \Eqref{eq:plrecur-deltat} to include $\Delta t$ requires us to reconsider the time reference updates for $k_t$. In line with the original definition, when the reset gate $r_t=0$, the time reference should remain the same. On the other hand, when $r_t=1$ we would like to update the time reference $k_t$ to be the current $t$. This poses a challenge for defining the forget gate correctly as $f_t \ne 0$ when $k_t=t$. We define then a separate time reference term $\tilde{k}_t$ as follows
\begin{equation}
    \tilde{k}_t = r_t \odot (t-\Delta t +1) + (1-r_t) \odot k_{t-\Delta t}.
    \label{eq:timeref_tilde}
\end{equation}
Plugging the time reference variables $k_t$ and \Eqref{eq:timeref_tilde} into \Eqref{eq:plrecur-deltat} and gives us the forget gate for the power law gated LSTM (pLSTM) with variable time points:
\begin{eqnarray}
f_{t} &=& \left(\frac{t-k_t+1}{t-\Delta t-\tilde{k}_t+1}\right)^{-p} \\
c_{t}&=& f_{t}\odot c_{t-\Delta t}+i_t\odot\tilde c_{t}. 
\end{eqnarray}
We note that the recurrent state on the previous time point $t-\Delta t$ is now given by $h_{t-\Delta t}$, $c_{t-\Delta t}$, and $k_{t-\Delta t}$ for the pLSTM. 

Further, the derivation in Section 3.3 of the gradient for the power law gate with variable time is similar in this case. The derivative of $f_t$ with respect to $r_t$ is given by
\begin{equation}
\frac{\partial f_t}{\partial r_t} = -p \cdot \frac{\left(c-c r_t\right)^{p-1}}{\left(a+ b r_t\right)^{p+1}} \cdot c(a+b),
\end{equation}
where $a=t-k_{t-\Delta t}+1$, $b=-t+k_{t-\Delta t}$, and $c=t-k_{t-\Delta t}+1-\Delta t$.
When $p\in (0,1)$ and the reset gate is activated ($r_t=1$), the derivative tends to infinity, i.e., $\frac{\partial f_t}{\partial r_t} \to \infty$. Similarly to our unit step version, we add a small $\varepsilon$ term in \Eqref{eq:forget_gate_plstm} such that $f_t=\left(\frac{t-k_t+1}{t-\Delta t-\tilde{k}_t+1+\varepsilon}\right)^{-p}$. 

When comparing performance in the frequency discrimination task to the update rule defined in \Eqref{eq:cell_state_plstm}, we show that the pLSTM learns time information even when it is not provided as input. Further in the asynchronous condition, we find that the model performance is better when relying on the timing information coming directly from the gate information as opposed to attending to this information as if it was input.

\subsection{Frequency Discrimination Task}

In the main text we introduce the Frequency Discrimination task \cite{neil2016phased} to understand how the time information is used by the power law forget gate. The task 
requires the use of time information to successfully learn to discriminate between two classes of sine waves.
One class is sine waves generated with a period in the range $T \sim \mathcal{U}(5,6)$ and the other class is sine waves generated with a period in the range $T \sim \{ \mathcal{U}(1,5) \cup \mathcal{U}(6,100)\}$, where $\mathcal{U}(a,b)$ indicates a uniform sample from the interval $(a,b)$. We tested three kinds of sequences in this classification task: (1) the synchronous condition where the points in sine waves are sampled with $\Delta t$ equal to 1 (sync-1); (2) another synchronous condition where the points in sine waves are over sampled with $\Delta t$ equal to 0.1 (sync-0.1); (3) the asynchronous condition where the points in sine waves are randomly sampled (async). We set all other parameters of the dataset (such as maximum duration) to those mentioned in the frequency discrimination task of \citet{neil2016phased}. We tested three models: LSTM, LSTM with chrono initialization (LSTM-chrono) and pLSTM. The two classes were balanced during creation of the dataset and all models were evaluated on their ability to distinguish between two classes. All models had 110 hidden units and 2 output units. The pLSTM uses an epsilon value of 1e-5 to ensure that epsilon is smaller than any observed time step across all sampling conditions.

We consider two conditions for this task regarding the input of the model. First, the input of the model only contains the amplitude of the points sampled from the sinusoidal sequence, $x$. Second, the input of the model contains a
pair $(x, t)$,  where $x$ is the amplitude and $t$ is the exact time information of the point sampled from the sinusoidal sequence. Because the forget gate is affected by the change in time (see \Eqref{eq:plrecur_supp}), it is possible that the pLSTM will learn information related to time directly from the gate without requiring $t$ to be considered input to the pLSTM. To test this, we compare performance of the LSTM variants when only the $x$ observation is provided to the network.

Note that in this experiment, we implemented the pLSTM with variable time points as described in \secref{sec:pLSTM_variable}, since the differences between the adjacent timesteps sampled from the sinusoidal sequences were not always 1. 

The results suggest that all models, regardless of the availability of the time information $t$, converged equally well in the sync-1 condition. In the sync-0.1 condition, we found that the pLSTM and LSTM-chrono both converged well and achieved higher accuracy than the LSTM, and this was in part because the LSTM did not always learn, with one of the 5 runs performing at chance. The most interesting result is the asynchronous condition in which the timestep between observations could be very small or quite large. We found that the LSTM and LSTM-chrono performed well only when the information of sampled time was available as part of the input. In contrast, while the pLSTM performed equally well as the LSTMs when the time information was available, the pLSTM performed significantly better than the LSTMs without the time information, in terms of converging rate and final accuracy (Table \ref{tab:FD}, Figure \ref{fig:fdt}). Altogether, the results suggest that the power law forget gate which contains time information as gate input can learn and maintain sequential information related to time. Access to this timing information at the gate level allows the pLSTM to have a better temporal precision outperforming other LSTM variants on the frequency discrimination task.
 

\subsection{Variable Copy Task}

In addition to the copy task with fixed T values as described in section 4.1, we also examined the models’ ability to capture the same task but with variable T values. We trained LSTM, LSTM-chrono and pLSTM models with sequences where T was randomly sampled from certain ranges: $T\sim \mathcal{U}([T_1, T_2])$, and validating the model performance with different sequences where T was also sampled from the same range. In the current experiment, we have two conditions: $T\sim \mathcal{U}([200, 500])$ and $T\sim \mathcal{U}([200, 1000])$. In both conditions, the training dataset contains 100000 sequences and the validation dataset contains 10000 sequences.

We found that LSTMs had difficulty learning the task in both conditions, possibly due to the fast decay of information. On the other hand, LSTM-chrono where we set $T_{max}=\frac{3}{2}T_2$, and pLSTM with uniform initialization of decay factor $p$, could learn the task and achieve convergence (Figure \ref{fig:copytask-varyT}A, \ref{fig:copytask-varyT}B)

We further examined how the pLSTM models captured the various copy task, by examining the models’ reset behavior. We found that the pLSTM model captured the task by using a small set of units as in the fixed T condition (section 5.2); furthermore, when processing different sequences where T were different, the model used the same set of units to encode information, by not resetting the reference time $k_t$ until the signal symbol showed up and the model needed to start reproducing the target symbols (Figure A.2C). This indicates that the pLSTM model learned to flexibly adjust the reset behavior of certain units to keep information within a flexible range.

\subsection{Generalization of Copy Task}

In this experiment, we examined the models’ ability to generalize their learned behavior in the copy task to the same task but with $T$ values they were not trained with. Specifically, we examined the generalization for models which were well-trained in the $T=200$ ($T_{200}$) and $T=500$ ($T_{500}$) conditions, by testing them with a range of $T$ values from $T=50$ to $T=2000$. We found that all models could generalize to different T values with an accuracy above chance (i.e. 1/9, Figure \ref{fig:copytask-generalization}). For the $T_{200}$ models, the LSTMs showed better generalization ability when $T$ is shorter than 200 compared to other models, while LSTM-chrono showed the better generalization ability when $T$ is longer than 200. The performance of pLSTMs fell in between. For the $T_{500}$ models, we found that pLSTM showed better performance when $T$ is shorter than 500, while LSTM-chrono and pLSTM showed comparable performance when generalizing to much longer sequences such as T=2000. 

The results suggest that LSTM and pLSTM models learned the task not by holding information for a certain amount of timesteps, but by identifying the signal symbol, and thus can adapt the behavior to tasks where the models need to hold information for a different amount of timesteps. Moreover, the result that LSTM-chrono generalized well for longer sequences but had difficulty generalizing to shorter sequences indicates the limitation of initializing the network to learn information in a certain timescale \emph{a priori}.

\renewcommand{\thefigure}{A.\arabic{figure}}
\begin{figure*}[ht]
\centerline{\includegraphics[width=\columnwidth*2]{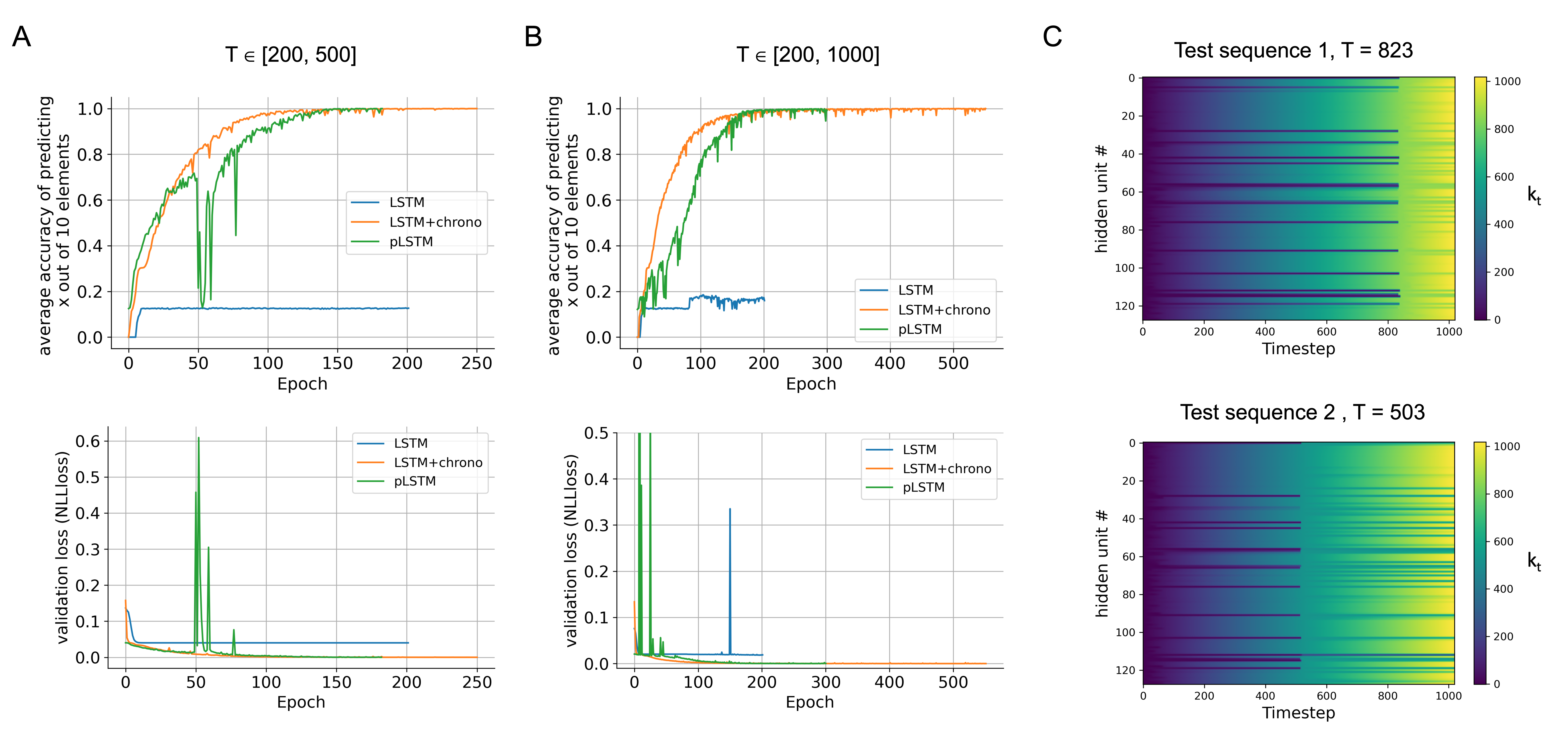}}
\caption{\textbf{Variable copy task results.} \textbf{A.} Variable copy task when T is randomly sampled from 200 to 500 timesteps. The results (measured by accuracy and validation loss) showed that both pLSTM and LSTM with chrono initialization $T_{max}=500$ could learn the task, while LSTM has difficulty learning the task. \textbf{B.} Variable copy task when T is randomly sampled from 200 to 1000 timesteps. Similarly, both pLSTM and LSTM with chrono initialization $T_{max}=1000$ successfully learned the task. \textbf{C.} Visualization of $k_t$ when pLSTM is processing two example test sequences with various T. The results suggest that by controlling the reset mechanism, pLSTM can flexibly keep information for various lengths and reproduce the target elements, without prior knowledge for initialization.}
\label{fig:copytask-varyT}
\end{figure*}

\begin{figure*}[ht]
\centerline{\includegraphics[width=\columnwidth*2]{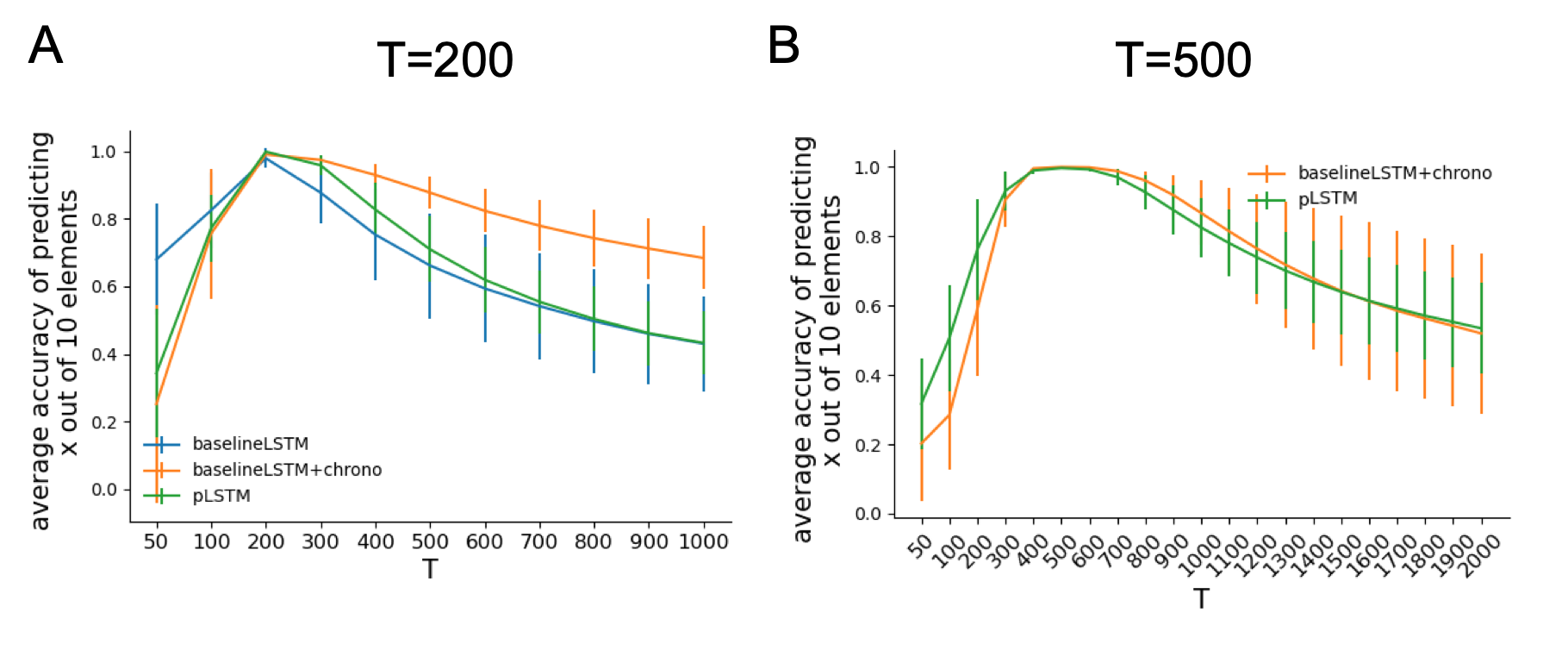}}
\caption{\textbf{Generalization of copy task}\textbf{A.} Average copy task accuracy when training models (5 models for each model type) with the T=200 condition and testing models with a range of T from T=50 to T=1000. All models could perform above chance when testing with sequences longer than the training sequences. LSTM showed advantage generalizing to shorter sequences while LSTM with chrono showed advantage generalizing to longer sequences. \textbf{B.} Average copy task accuracy when training models (5 models for each model type) with the T=200 condition and testing models with a range of T from T=50 to T=2000. pLSTM showed advantage for generalizing to shorter sequences.}
\label{fig:copytask-generalization}
\end{figure*}

\begin{figure*}[ht]
\centerline{\includegraphics[width=\columnwidth*2]{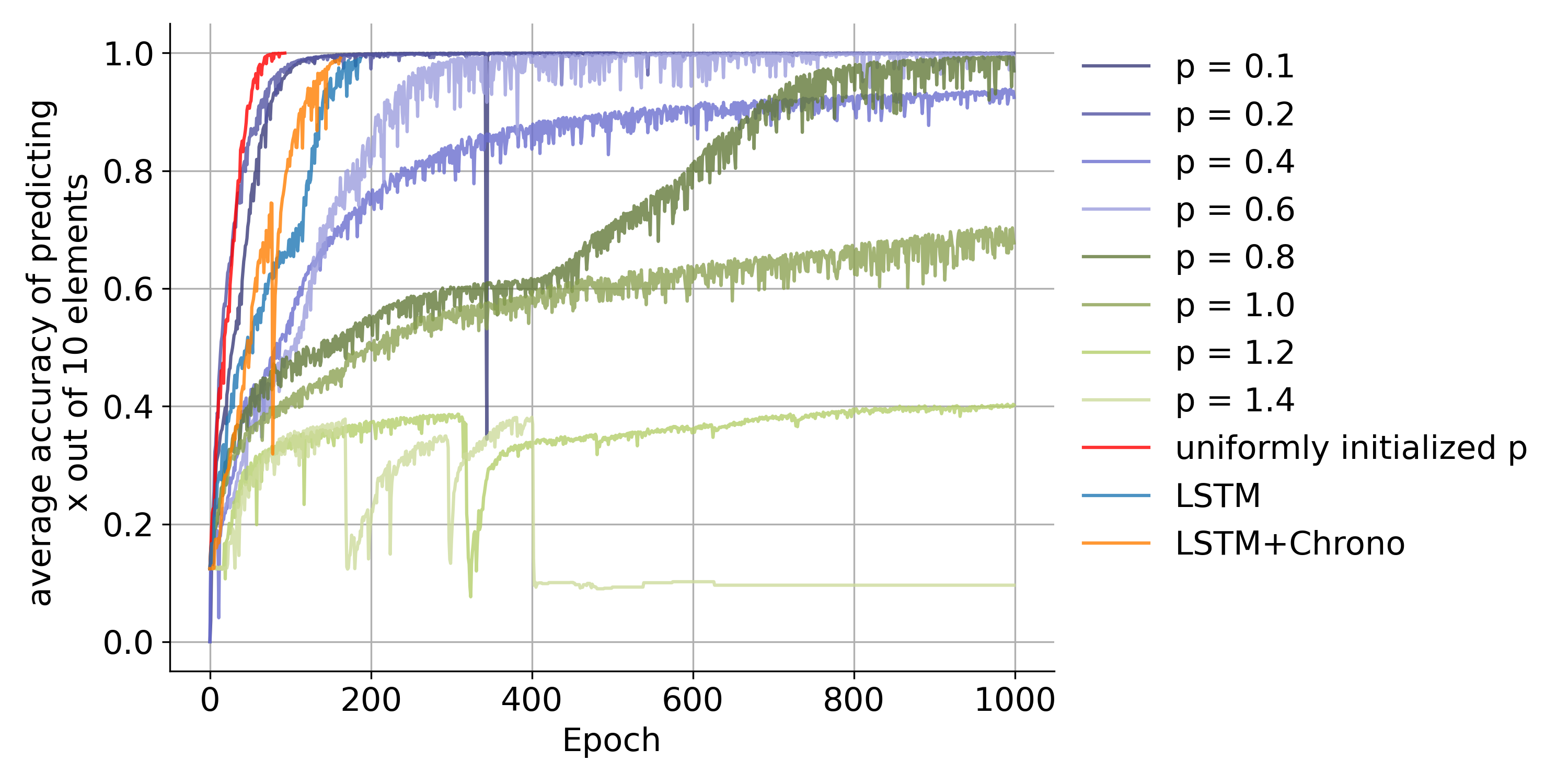}}
\caption{\textbf{Copy task (T=200) accuracy when fixing decay factor $p$ to different values.} The power law gate controls information decay following the decay factor $p$ in the power law function. As expected, smaller $p$ indicates a slower decay of information, and thus yields a faster convergence of copy task which requires the model to keep information over hundreds of timesteps. pLSTMs with $p<1$ successfully learned the task, following the pattern that smaller $p$ learned faster. pLSTMs with $p>1$ have difficulty learning the task.}
\label{fig:copytask-varyp}
\end{figure*}

\begin{figure*}[ht]
\centerline{\includegraphics[width=\columnwidth*2]{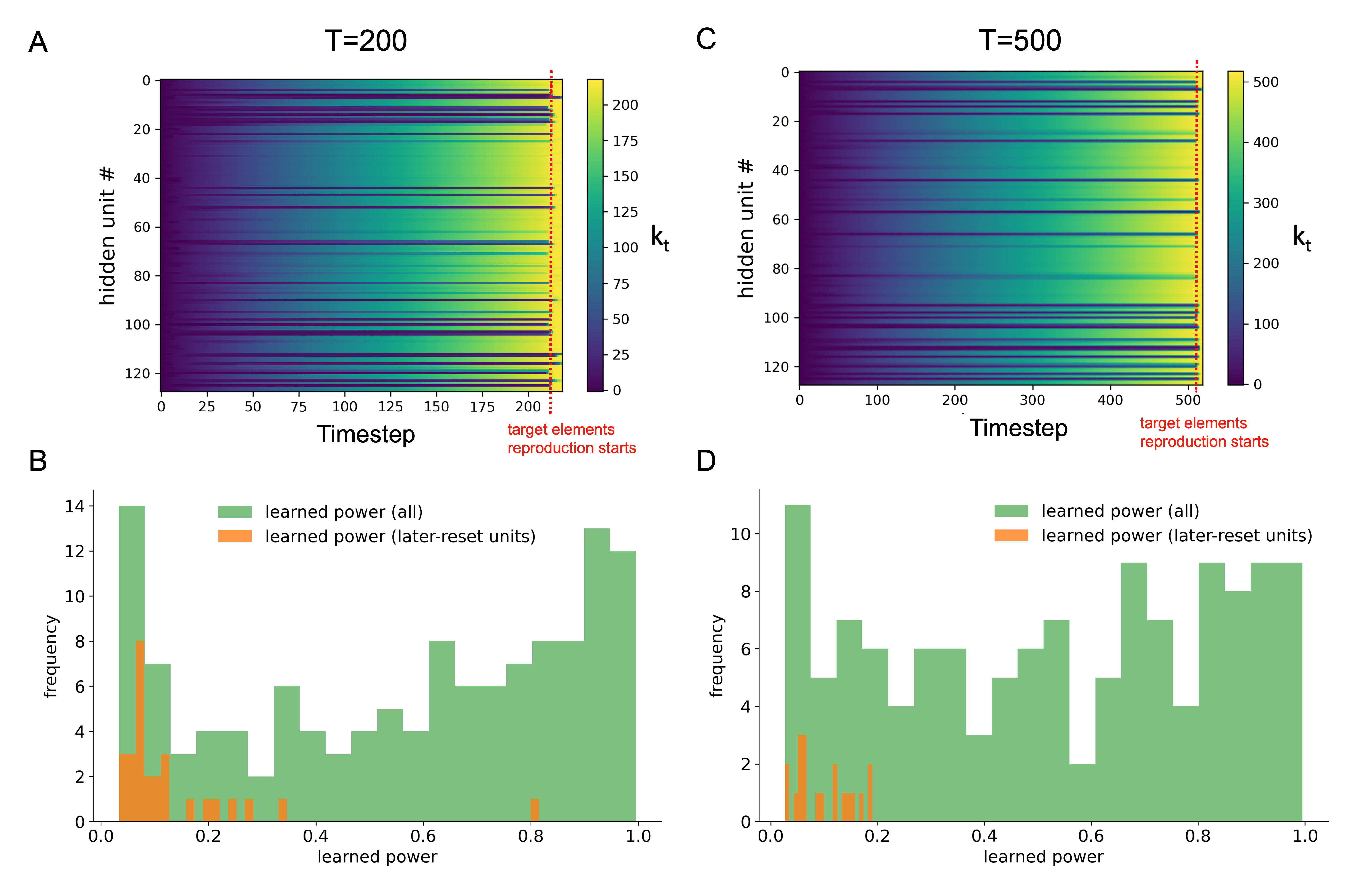}}
\caption{\textbf{Reference time $k_t$ in pLSTM during the copy task, and the $p$ distribution after learning the copy task} \textbf{A.} $k_t$ during copy task (T=200). Most units were constantly resetting the reference time $k_t$ to follow the current timestep, while $\sim$25 units (the ``later-reset units") only started to reset the reference time at the moment that target elements reproduction starts, or later. This suggests a small subset of units in pLSTM were keeping information for achieving the copy task. \textbf{B.} $p$ distribution after learning the copy task (T=200). The $p$ was uniformly initialized between 0 and 1. After learning the copy task, the $p$ distribution slightly shift toward right. The later-reset units, however, are units with relatively smaller $p$. \textbf{C.} $k_t$ during copy task (T=500). Similar to T=200 condition, most units were constantly resetting at every timestep, except for a small subset of units ($\sim$20 units) which were resetting later after the reproduction starts.\textbf{D.} $p$ distribution after learning the copy task (T=500).} Compared to the T=200 condition, the $p$ distribution in the T=500 condition contains more values that are smaller than 0.5, indicating that the network adjusts the decay factor to learn task-relevant timescales.
\label{fig:pdist}
\end{figure*}

\begin{figure*}[ht]
\begin{center}
 \begin{minipage}{0.49\linewidth}
  \centerline{\includegraphics[width=\columnwidth]{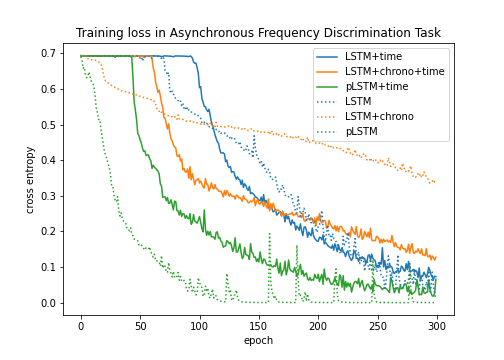}}
 \end{minipage}
\hfill
 \begin{minipage}{0.49\linewidth}
  \centerline{  \includegraphics[width=\columnwidth]{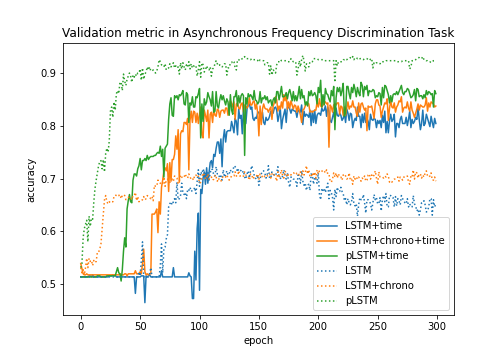}}
 \end{minipage}
\end{center}
\caption{\textbf{Training loss and validation accuracy in asynchronous frequency condition.} The power law gate includes information related to time and thus may improve performance on time-dependent tasks. We show the training loss and validation metric for a single run of each model type tested in the frequency discrimination task. The sequential samples of data occur at random timesteps.}
\label{fig:fdt}
\end{figure*}

\end{document}